# On Appropriate Selection of Fuzzy Aggregation Operators in Medical Decision Support System

K.M. Motahar Hossain, Zahir Raihan, M.M.A. Hashem
Department of Computer Science and Engineering,
Khulna University of Engineering & Technology (KUET), Khulna-920300, Bangladesh.
shibly99@hotmail.com, raihan306@yahoo.com, hashem@cse.kuet.ac.bd

**Abstract**

*The Decision Support System (DSS) contains more than one antecedent and the degrees of strength of the antecedents need to be combined to determine the overall strength of the rule consequent. The membership values of the linguistic variables in Fuzzy have to be combined using an aggregation operator. But it is not feasible to predefine the form of aggregation operators in decision-making. Instead, each rule should be found based on the feeling of the experts and on their actual decision pattern over the set of typical examples. Thus this work illustrates how the choice of aggregation operators is intended to mimic human decision making and can be selected and adjusted to fit empirical data – a series of test cases. Both parametrized and non-parametrized aggregation operators are adapted to fit empirical data. Moreover, they provided compensatory properties and, therefore, seemed to produce a better decision support system. To solve the problem, a threshold point from the output of the aggregation operators is chosen as the separation point between two classes. The best-achieved accuracy is chosen as the appropriate aggregation operator. Thus a medical decision can be generated, which is very close to a practitioner's guideline.*

## I. INTRODUCTION

Expert and decision support systems are common in the areas of machine learning where the alternatives are selected based on combined support of a number of factors, none of which could determine the alternative by itself. An example of such an area is medicine, where diagnosis or management is almost never decided based on individual criterion. A weighted combination of many criteria is used instead; each criterion may support various alternatives, and the alternative with the strongest support is selected as the decision. Fuzzy systems are being used successfully in an increasing number of application areas of these types; they use linguistic rules to describe systems. One of the most important considerations in designing any fuzzy system is the generation of the fuzzy rules as well as the membership functions for each fuzzy set. In most existing applications, the fuzzy rules are generated by experts in the area, especially for problems with only a few inputs.

With an increasing number of variables, the possible number of rules for the system increases exponentially, which makes it difficult for experts to define a complete rule for good system performance. An automated way to design fuzzy systems might be preferable [1].

The aims of this work are twofold: First is to discuss various methods of selecting and adjusting aggregation operators based on empirical data and expert opinion. We consider the situations where no priori knowledge about the properties of operators is available, and therefore, they have to be built using exclusively empirical data, as well as the situations where the expert opinion dictates the form of the operator and the available free parameters are adjusted to fit the data. Second is to illustrate this process on some standard medical datasets.

In this Medical DSS, some common aggregation operators are selected. Every dataset contains real data of various pathological test results for huge number of patients collected in different times. For each dataset all the selected operators are applied simultaneously to the empirical data and the generated output is compared with the actual output to test the accuracy. For different operators the calculated accuracy differs for any particular medical dataset. So, same operator did not show satisfactory results all the time for different datasets. This work selected 14 various operators from different classes to make a wide selection domain. Thus the decision support system can generates more accurate results with this wide range of selection variables.

## II. STATE OF THE ART

### A. Decision Making in Fuzzy Environment

In classical (normative, statistical) decision theory a decision can be characterized by a set of decision alternatives (the decision space); a set of states of nature (the state space); a relation assigning to each pair of a decision and state a result; and finally, the utility function which orders the results according to their desirability. Multi criteria decision making has led to numerous evaluation schemes (e.g. in the areas of cost benefit analysis and marketing) and to the formulation of vector-maximum problems in mathematical programming [7].

A part from this, two major areas have evolved, both of which concentrate on decision making with several criteria: Multi Objective Decision Making (MODM) and Multi Attribute Decision Making (MADM).

The general multi attribute decision-making model can be defined as follows:

Let $X = \{x_i \mid i = 1, \ldots, n\}$ be a (finite) set of decision alternatives and $G = \{g_j \mid j = 1, \ldots, m\}$ a (finite) set of goals according to which the desirability of an action is judged. Determine the optimal alternative $x^0$ with the highest degree of desirability with respect to all relevant goals $g_j$. Most approaches in MADM consist of two stages:

1. The aggregation of the judgments with respect to all goals and per decision alternative.

2. The rank ordering of the decision alternatives according to the aggregated judgments.

The problem under consideration is a typical one of multi criteria decision making (MCDM), various approaches to which have been discussed in [3], [7] and [10]. One important class of methods in MCDM is based on constructing a utility or value function $u(x)$, which represents the overall strength of support in favor of the alternative $x$. This approach is known as multi attribute utility theory (MAUT). In MAUT, one can represent the preference relation $\succeq$ on a set of alternatives $X$ with a single-valued function $u(x)$ on $X$, called utility, such that for any $x, y \in X, x \succeq y \Leftrightarrow u(x) \geq u(y)$. Maximization of $u(x)$ over $X$ provides the solution to the problem of selecting $x$. The conditions and criteria for the utility functions are explained in [1] and [10].

### III. PRELIMINARIES

In this section, we review some aggregation operators that are used later on in this work. An aggregation operator is simply a function, which assigns a real number $y$ to any n-tuple $(x_1, x_2, \ldots, x_n)$ of real numbers:

$y = Agg(x_1, x_2, \ldots, x_n)$  (1)

Now, the choice of aggregation operator in decision support system is considered. The vector maximization problem of MAUT [1],

Maximize $(u_1(x_1), u_2(x_2) \ldots \ldots u_n(x_n))$ over $X$

where $u_i(x_i)$ are the utility functions of the corresponding attributes $x_i$, takes the form

Maximize $u(x) = U(u_1(x_1), u_2(x_2) \ldots \ldots u_n(x_n))$  (2)

where $U$ is some function of n real variables which aggregates the individual utility values into the overall utility values into the overall utility of the alternative x. In FST this problem takes form

Maximize $\mu_\Omega(x) = Agg(\mu_{A1}(x_1), \mu_{A2}(x_2), \ldots, \mu_{An}(x_n))$  (3)

Where, $Agg$ stands for an appropriate aggregation operator. It combines the membership values in the sets $A_1, A_2 \ldots A_n$, into the membership value of the set $\Omega$, formed by some operation on the sets $A_1, A_2 \ldots A_n$, such as intersection, union, or their combination.

There are two basic classes of operators: operators for the intersection and union of fuzzy sets- referred to as triangular norms and conorms and the class of averaging operators, which model connectives for fuzzy sets between t-norms and t-conorms. Each class contains parametrized as well as nonparametrized operators [4], [7].

A straight forward approach for aggregating fuzzy sets, for instance in the context of decision making, would be to use the aggregating procedures frequently used in utility theory or multi criteria decision theory. They realize the idea of trade-offs between conflicting goals when compensation is allowed and the resulting trade-offs lie between the most optimistic lower bound and the most pessimistic upper bound, that is, they map between the minimum and maximum degree of membership of the aggregated sets. Therefore, they are called the averaging operators. Operators such as the weighted and unweighted arithmetic or geometric mean are examples of nonparametric averaging operators. In fact, they are adequate models for human aggregation procedures in decision environments and have empirically performed quite well [7]. Among these The compensatory and is more general in the sense that the compensation between intersection and union is expressed by a parameter $\gamma$. The $\gamma$ operator is a combination of the algebraic product and the algebraic sum. There is a special relationship between parametrized families of operators and the $t$-norms & $t$-conorms with respect to special values of their parameters [7].

### IV. PROBLEM FORMULATION & STATEMENT

Consider the problem of fitting an aggregation operator $f(x_1, x_2, \ldots, x_n)$ to the empirical data. The empirical data consists of a list of pairs/triples/n tuples of membership values to be aggregated $\{(x_1^k, x_2^k, \ldots, x_n^k)\}_{k=1}^{K}$ and the corresponding compound membership values $\{d^k\}_{k=1}^{K}$, measured experimentally.

The aggregation operator is an n-place function $f : [0,1]^k \to [0,1]$, non-decreasing in all arguments and satisfying $f(0) = 0$ and $f(1) = 1$. This function also is called general aggregation operator and the foregoing properties constitute the minimal set of properties aggregation operators must satisfy. The continuity of f is also required, because this property is important from a practical point of view.

Additional properties define particular classes of aggregation operators. For example, commutativity, associativity, and boundary condition $f(x,0) = 0$ define the well-known class of triangular norms.

The problem is formulated as follows [9]:

minimize $\|f(x_1, x_2, \ldots, x_n) - d\|$,

subject to $f$ belonging to a given class of aggregation operators. $d$ denotes the $K$ vector of measured compound membership values.

It is noted from the beginning that generally this is an approximation and not an interpolation problem; i.e., f needs not fit the empirical data exactly, and the data itself may not satisfy the properties required from the operator (e.g., data points may violate commutativity condition; hence, no commutative operator can fit the data exactly). It is clear that empirical data may contain some measurement errors, missing attributes and hence an operator is searched that approximates the data.

Given empirical data $\{(x_i, y_i)\}_{i=1}^{N}$, and possibly experts' opinion about the aggregation operator, find such representation of the aggregation operator that

1) Provides good approximation to empirical data;
2) Is flexible to model various classes of aggregation operators;
3) Is able to confine to a particular class of operators, or a particular property;
4) Is semantically clear.

That is, some empirical data in the form of datasets of various diseases are given including the class attribute of those datasets. The class attribute indicates the actual outcome of the data. Each instance of the dataset has its actual result or outcome. It can also be expressed as the Expert's opinion about each instance of the diseases. Each instance has N-1 number of attributes which indicates the symptoms or factors of that particular disease. The value of those attributes causes the class attribute to fall into a particular class. Thus the value of the class attribute depends only upon the value of N-1 attributes.

In this problem, some aggregation operators are chosen and those operators are applied to the empirical data. The output value of each aggregation operator causes the result to be any particular class. Then the measured value is compared with the given class attribute value to get the accuracy of that particular aggregation operator for that particular dataset. By concurrently applying a number of chosen aggregation operators to a particular dataset, operator with a highest degree of accuracy is chosen as the operator for that very dataset.

## V. FITTING EMPIRICAL DATA TO THE AGGREGATION OPERATORS

The whole process of the Medical Decision Support System is illustrated here. It contains the choosing of membership functions in fuzzification process and selecting of various parameters of the aggregation operators, combine the results of the aggregation operators, selection of the threshold points between classes, and classify the result according to the threshold point all include the defuzzification process.

### A. Selection of Membership Function

Each attribute contains data that are either nominal or ordinal depends upon the type of the attribute. Thus the membership function used is also depends on it [7].

*Nominal attribute value:* If the attribute value is nominal then the sigmoid membership function is chosen. Other membership function such as left triangular, right triangular or Z– membership function [7] can also be used. But for the smoothness and simplicity of calculation, sigmoid membership function is proposed here. Point to be noted that, in this work the highest membership value must be less than 1 and the lowest membership value must be greater than 0. That is,
$$0 \leq \mu(x) \leq 1$$

Because the class of operators used here are mostly t-norms, t-conorms and averaging operators. The min, product, and bounded sum operators belong to the so called t-norms class and the max-operator, algebraic sum and bounded sum belongs to the t-conorms class. Thus the membership value of 0 or 1 may cause the result of the aggregation operator either 0 or 1, which may cause to be biased.

*Ordinal attribute value:* If the attribute value is ordinal then either the sigmoid or Gaussian or left triangular, right triangular, triangular or Z –membership function may used, depends upon the data.

### B. Fuzzification and Data Normalization

After choosing the membership value of each attribute, the data must be fuzzified using each membership function. Here, the missing attribute and the ordering of the attribute must be taken into account. It contains some steps;

Step 1: The class attribute should be placed as the last attribute among all.

Step 2: The nominal attribute is converted into 0 or 1. The highest value is replaced by 1 and the lowest value is replaced by 0.

Step 3: The missing value of any attribute is replaced with the mean value of that particular attribute. In the case of nominal attribute the mean value is rounded into either 0 or 1.

After these steps the fuzzification process is started. In this process, the center and width of each attribute is calculated. Various procedures are applicable for finding the center and width of attribute for different membership function. In this work Gaussian and sigmoid membership functions are used in most of the attributes of datasets.

$$f_{Gaussian}(x) = e^{-0.5y^2} \quad \text{where} \quad y = \frac{8(x - x_1)}{x_2 - x_1} - 4 \qquad (4)$$

$$f_{Sigmoid}(x) = \frac{1}{1 + e^{(-y+6)}} \quad \text{where } y = \frac{12(x - x_1)}{x_2 - x_1} \qquad (5)$$

Example 1: To analyze the feasibility of the approach, four artificial problems have been analyzed. These problems have been generated using the publicly available information from machine learning repository [2].

1. Breast-cancer-Wisconsin dataset (11 attributes and 699 examples).
2. Hepatitis dataset (20 attributes and 155 examples).
3. Lymphography dataset (19 attributes and 148 examples).
4. Echocardiogram dataset (13 attributes and 132 examples).

Some dataset contains some missing values such as Breast-cancer-Wisconsin dataset has 16 missing values in 16 instances. Only the Lymphography dataset contains no missing values. Here is an example of 10 instances from lymphography dataset and as well as the fuzzified data are shown in the Table I and Table II:

Table I: Instances from Lymphography datasets

| Lymphat-ics | Block of affere | Bl. of lymph. c | …… | No. of nodes in | Class value |
|---|---|---|---|---|---|
| 2 | 2 | 1 | …… | 1 | 2 |
| 3 | 2 | 2 | …… | 7 | 2 |
| 1 | 1 | 1 | …… | 2 | 1 |
| 2 | 1 | 1 | …… | 6 | 3 |

Table II: Fuzzified data from Lymphography data instances

| Lymphat-ics | Block of affere | Bl. of lymph. C | … | No. of nodes in | Class value |
|---|---|---|---|---|---|
| 0.606531 | 0.606531 | 0.135335 | … | 0.52045 | 0.4 |
| 0.800737 | 0.606531 | 0.606531 | … | 0.960005 | 0.4 |
| 0.411112 | 0.135335 | 0.135335 | … | 0.606531 | 0.2 |
| 0.606531 | 0.135335 | 0.135335 | … | 0.912254 | 0.6 |

### C. Application of Aggregation Operators and Its Properties

In this step number of aggregation operators of some predefined classes is applied concurrently. In this work, various operators of t-norms, t-conorms and averaging classes are used. These classes of aggregation operator possess some special properties. *t-norms* are two-valued functions t from $[0,1] \times [0,1]$ which satisfy monotonicity, commutativity and associativity [4],[6],[7]. *t-conorms* or *s-norms* are also associative, commutative, and monotonic two placed functions s, which map from $[0,1] \times [0,1]$ into $[0,1]$. So, t-norms and t-conorms are related in a sense of logical duality. t-conorm can be defined as a two-placed function s mapping from $[0,1] \times [0,1]$ in $[0,1]$ such that the function t, defined as

$$t(\mu_A(x), \mu_B(x)) = 1 - s(1 - \mu_A(x), 1 - \mu_B(x)) \quad (6)$$

is a t-norm. So any t-conorm s can be generated from a t-norm through this transformation.

Thus they can be used in N number of attributes in the datasets. The result of each aggregation operator for each instance gives a real number. This is further compared with the expert's opinion or the actual outcome.

### D. Finding the Threshold Point between Two Classes

Procedure for calculating the threshold point for any particular aggregation operator is given below:

Proc_cal_t_point(ok, nok, fnok, fok, f_m_point, step_size, compare, m_point, n) /* ok & nok are indicators and fok & fnok are final value of the corresponding indicators. ok is used when the calculated value matches the actual value and nok is used when it fails. m_point is the lower threshold point and compare is the upper threshold point for the overlapping value of aggregated value which fall into more than one class. f_m_point is the final value between two adjacent classes by which these classes can be classified mostly. n is the number of total instances.*/

1. **while** gama>-1 /* Here gama is the parameter of an aggregation operator. By changing the value of the gama, the parameterized Aggregation operator can fall into a particular operator. */
2.     t_m_point ← m_point /* temporary mid point or t_m_point is used for calculation. */
3.     **while** t_m_point<=compare
4.        **for** each i ∈ n /* for $i^{th}$ instance of dataset*/
5.           **read** temp[i]
6.           result ← agg_op(temp[i], N) /* N is the number of attributes in an instance. Here the agg_op function gets an instance of the dataset as an input and put the aggregated output into the result.*/
7.           **if**(temp[i][N]=K & result <= t_m_point) /*it checks whether for a particular value (K) of the result instance ($N^{th}$ attribute), the aggregated result value crosses the temporary threshold point or not*/
8.           **then** ok + 1
9.           **else** nok+ 1 /* these indictors increases according to the above condition. ok indicates the frequency of the correct match and nok indicated the frequency for incorrect match. */
10.        **end for**
11.     **if**(nok<fnok) /*this condition is used to minimize the incorrect match. */
12.        **then** fnok ← nok, fok ← ok, f_m_point ← t_m_point
13.     ok ← 0, nok ← 0
14.     t_m_point ← t_m_point + step_size /* By changing the value of the temporary threshold point /mid point by increasing with a regular interval which is defined as step_size, the frequency of the incorrect match can be minimized. When the minimum incorrect match is found, then it terminates. */
15.     **end while**
16.     gama=gama-1
17. **end while**

Each threshold point indicates a linear divider between two consecutive classes. Thus after finding the threshold point, the result of the operator can be classified to any one of the classes considering the threshold points between which it belongs to. When the threshold point for each aggregation operator for any given dataset is calculated then it is an easy task to calculate the accuracy for that operator by comparing the classified result with the given result.

### E. Selecting the Appropriate Aggregation Operator

The operator with the highest degree of accuracy is considered to be the operator for that particular dataset. Thus that calculated threshold point using the selected aggregation operator could classify any new instances of that dataset. Point to be noted that, if more than one aggregation operator perform the same highest accuracy, the operator with a maximum number of classified instances is chosen to be appropriate.

## VI. EXPERIMENTAL RESULTS

In this work three classes of aggregation operators such as: t-norms, t-conorms and averaging operators are used. Each of which has a number of operators, which are shown in Table III:

Table III: Used Aggregation Operators according to classes

| Aggregation Operators of T-norms Class | Aggregation Operators of T-conorms Class | Averaging operators |
|---|---|---|
| Einstein Product | Einstein Sum | Fuzzy AND operator |
| Algebric Product | Algebric Sum | Fuzzy OR operator |
| Hamacher Product | Hamacher Sum | Convex combination of min |
| MIN Operator | MAX Operator | |
| Dombi Intersection | MAX Operator | Convex combination of max |

For t-norms and t-conorms only Hamacher, Dubois, Dombi (both Union and Intersection) are used as they are parametrized operators. Only changing the parameter can change the class and operator. Here are the results that are generated by applying the above-mentioned methodologies to those four standard datasets.

Table IV: Result of Breast-cancer-Wisconsin Dataset

| Operator Class name | Aggregation Operator name | Threshold Point | Correct Instance | Accuracy in % |
|---|---|---|---|---|
| T-norms | **Einstein Product** | **1.10276e-05** | **682** | **97.568** |
| | Algebric Product | 0.00058323 | 681 | 97.4249 |
| | Hamacher Product | 2.88 | 678 | 96.9957 |
| | MIN Operator | 10 | 537 | 76.824 |
| | Dombi Intersection | 5 | 241 | 34.4778 |
| T-conorms | Einstein Sum | 99 | 677 | 96.8526 |
| | Algebric Sum | 96 | 676 | 96.7096 |
| | Hamacher Sum | 80 | 650 | 92.99 |
| | MAX Operator | 10 | 537 | 76.824 |
| | Dombi Union | 62 | 458 | 65.5222 |
| Averaging Operator | Fuzzy AND | 11.18 | 646 | 92.4177 |
| | Fuzzy OR | 45.0484 | 669 | 95.7082 |
| | Convex combination of min | 0.19 | 667 | 95.422 |
| | Convex combination of max | 0.3 | 663 | 94.8498 |

Table V: Result of Hepatitis Dataset

| Operator Class name | Aggregation Operator name | Threshold Point | Correct Instances | Accuracy in % |
|---|---|---|---|---|
| T-norms | **Einstein Prod.** | **0.0049001** | **124** | **80** |
| | **Algebric Prod.** | **0.100601** | **124** | **80** |
| | **Hamacher Prod** | **6.78006** | **124** | **80** |
| | MIN Operator | 33 | 123 | 79.3548 |
| | Dombi Int. | 12 | 32 | 20.6452 |
| T-conorms | Einstein Sum | 50 | 123 | 79.3548 |
| | Algebric Sum | 50 | 123 | 79.3548 |
| | Hamacher Sum | 50 | 123 | 79.3548 |
| | Dubois Union in max | 1 | 123 | 79.3548 |
| | Dombi Union | Non satisfactory result | | |
| Averaging Operator | Fuzzy AND | 39 | 123 | 79.3548 |
| | Fuzzy OR | 62 | 123 | 79.3548 |
| | Convex combination of min | 0.4 | 123 | 79.3548 |
| | Convex combination of max | 0.8 | 123 | 79.3548 |

Table VI: Result of Lymphography Dataset

| Operator Class name | Aggregation Operator name | Threshold Point | Correct Instance | Accuracy in % |
|---|---|---|---|---|
| T-norms | Einstein Product | 1e-12, 0.00100001, 1e-08 | 121 | 81.7568 |
| | **Algebric Product** | **1.001e-09, 0.015, 1e-08** | **144** | **97.2973** |
| | **Hamacher Prod.** | **1.4, 6.1, 1.6** | **144** | **97.2973** |
| | MIN Operator | | | |
| | Dombi Intersection | | | |
| T-conorms | Einstein Sum | Non satisfactory result | | |
| | Algebric Sum | | | |
| | Hamacher Sum | | | |
| | MAX Operator | | | |
| | Dombi Union | | | |

| Operator Class | Aggregation Operator name | Threshold Point | Correct Instances | Accuracy in % |
|---|---|---|---|---|
| Averaging Operator | Fuzzy AND | 24, 36, 62 | 75 | 50.6757 |
| | Fuzzy OR | 49, 76, 93 | 106 | 71.6216 |
| | Convex combination of min | 0.37, 0.82, 0.62 | 125 | 84.4595 |
| | Convex combination of max | 0.39, 0.93, 0.75 | 112 | 75.6757 |

Table VII: Result of Echocardiogram Dataset

| Operator Class name | Aggregation Operator name | Threshold Point | Correct Instances | Accuracy in % |
|---|---|---|---|---|
| T-norms | Einstein Product | 33.67 | 105 | 79.5455 |
| | Algebric Product | 43.4284 | 105 | 79.5455 |
| | Hamacher Product | 51.4872 | 105 | 79.5455 |
| | MIN Operator | 64.8178 | 94 | 71.2121 |
| | Dombi Intersection | 53 | 88 | 66.6667 |
| T-conorms | Einstein Sum | 99 | 88 | 66.6667 |
| | Algebric Sum | 99 | 88 | 66.6667 |
| | Hamacher Sum | Non satisfactory result | | |
| | MAX Operator | 88.2518 | 88 | 66.6667 |
| | Dombi Union | 98 | 44 | 33.3333 |
| Averaging Operator | Fuzzy AND | 95.7954 | 87 | 65.9091 |
| | Fuzzy OR | 97.6964 | 93 | 70.4545 |
| | Convex combination of min | 0.953609 | 96 | 72.7273 |
| | **Convex combination of max** | **0.92252** | **115** | **87.1212** |

Fourteen different operators of three predefined classes are used which are discussed previously. These operators show different accuracy for different datasets. For each dataset the Aggregation Operator, which is highlighted, that shows the highest accuracy is chosen. Table VIII gives a comparison of obtained accuracies with those of past works[2].

Table VIII: Comparison of accuracies with those of previous works in the following datasets

| Breast-Cancer-Wisc. | Hepatitis | Lymphography | Echocardiogram |
|---|---|---|---|
| Fuzzy DSS → 97.57% | Fuzzy DSS → 80% | Fuzzy DSS → 97.30% | Fuzzy DSS → 87.12% |
| Pattern Separation(1990) → 95.90% | Knowledge-Elicitation Tool(1987) → 83.00% | Induction in Noisy Domains (1987) → 83.00% | Exemplar-Based Learning (1988) → 75.76% |
| Instance-Based Learning (1992) → 93.70% | Computer-Int Methods (1983) → 80.00% | Knowledge-Elicitation Tool(1987) → 76.00% | Productive Value (1986) → 61.00% |

## VII. CONCLUSION

The overall goal of this work was to develop and analyze a new technique of decision support system, which can be applied successfully to solve different kinds of problems of medical diagnosis. The major application of the proposed method may be in decision support systems, where little is known about how the criteria (risk factors, indicators, etc.) should be aggregated. Aggregation operators were used in this system for combining and generating the output into a single form, as they have some special properties. Thus for different datasets very satisfactory results for different operators were achieved. The scope of the problem is limited to the linier classifier and selection of membership function is restricted among the traditional membership functions like sigmoid, Gaussian etc. Customize choice of membership function may also increase the accuracy of the result. Such computerized clinical guidelines can provide significant benefits to health outcomes and costs if they are implemented effectively.